\pdfoutput=1
\documentclass[10pt,twocolumn,letterpaper]{article}

\usepackage[pagenumbers]{cvpr} 

\usepackage{cuted}      
\usepackage{caption}
\usepackage{multirow}
\usepackage{amsthm}
\usepackage{bm}

\newtheorem{proposition}{Proposition}

\newcommand{\Vinfo}{$\mathcal{V}$-information\xspace}
\newcommand{\VInfo}{$\mathcal{V}$-Information\xspace}
\newcommand{\Cc}{\mathcal{C}}
\newcommand{\Cg}{\mathcal{C}_g}
\newcommand{\Gc}{\mathcal{G}}
\newcommand{\Yc}{\mathcal{Y}}
\newcommand{\E}{\mathbb{E}}
\newcommand{\method}{CIEG\xspace}
\newcommand{\Method}{CIEG\xspace}
\newcommand{\methodfull}{context-independent evidence governance\xspace}
\newcommand{\Methodfull}{Context-independent evidence governance\xspace}
\newcommand{\sig}{$^\dagger$}
\newcommand{\best}[1]{\textbf{#1}}

\usepackage{colortbl}
\definecolor{oursblue}{HTML}{0072B2}
\definecolor{imgteal}{HTML}{00897B}
\definecolor{grpgray}{HTML}{F3F4F6}

\definecolor{cvprblue}{rgb}{0.21,0.49,0.74}
\usepackage[pagebackref,breaklinks,colorlinks,allcolors=cvprblue]{hyperref}

\def\confName{CVPR}
\def\confYear{2027}

\title{Not All Evidence Is Admissible:\\Evidence Governance for Tool-Augmented Medical Agents}

\author{
Runxun Zhang$^{1}$ \quad
Onofrio Antonio Catalano$^{2}$ \quad
Jinqiao Wang$^{3}$ \quad
Jingwei Wei$^{1}$\\[4pt]
$^{1}$Institute of Automation, Chinese Academy of Sciences \quad
$^{2}$Massachusetts General Hospital\\
$^{3}$University of Chinese Academy of Sciences
}

\begin{document}
\maketitle
\begin{strip}
\centering
\makebox[0.695\linewidth][l]{\small\textbf{(a)}}\hfill\makebox[0.281\linewidth][l]{\small\textbf{(b)}}\par\vspace{1pt}
\includegraphics[width=0.695\linewidth]{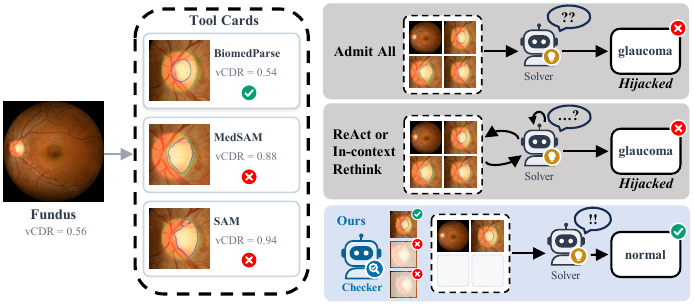}\hfill
\includegraphics[width=0.281\linewidth]{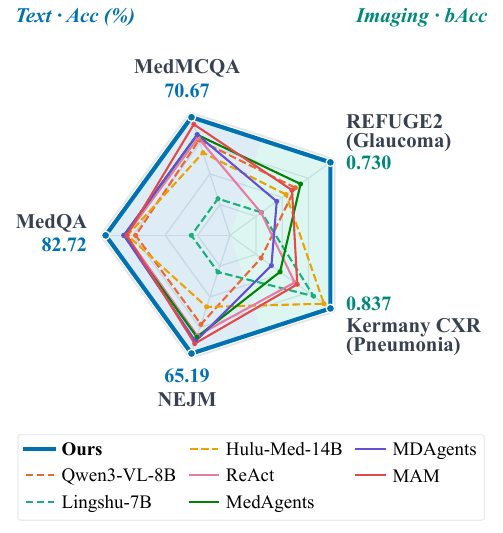}
\captionsetup{hypcap=false}
\captionof{figure}{\textbf{Evidence hijacking and its remedy.}
\textbf{(a)}~For a normal fundus image, admitting all biased tool cards flips the Solver to a false diagnosis. Contextual reasoning (e.g., ReAct) also suffers from this hijacking. Our Checker filters cards independently and restores the correct prediction.
\textbf{(b)}~Main results: accuracy (\%) on MedQA, MedMCQA, and NEJM, and bAcc on REFUGE2 and Kermany pneumonia classification. Ours (\method) is outermost on all five axes.}
\captionsetup{hypcap=true}
\label{fig:teaser}
\end{strip}

\begin{abstract}
Tool-augmented medical agents treat whatever their tools return as evidence. We show that this default is costly: medical tools fail silently, and a frozen reasoner cannot ignore what enters its context, so many cases it would have answered correctly are flipped to wrong once tool outputs are added. We call this \emph{evidence hijacking}. \VInfo explains the phenomenon: the reasoner has lost the option to ignore, so a tool output can carry negative usable information, and moving the decision of what to keep outside the reasoner restores that option. The theory further shows that part of any filtering gain comes merely from letting the reasoner see \emph{less}, and only selection by content lets it see \emph{better}. Building on this, we recast tool-augmented reasoning as \emph{evidence governance}, which separates deciding what counts as evidence from the reasoner that answers, and propose \emph{\methodfull} (\method), the first medical agent framework that rules, after acquisition, on whether each individual tool output may enter as evidence. Every tool output is written as a card that can be checked against the raw input; an independent checker that sees only the input and the cards decides which cards become evidence; and the reasoner answers from the admitted cards alone, with no training anywhere. On medical question answering, fundus glaucoma screening, and chest X-ray pneumonia classification, it achieves the highest accuracy among models of comparable scale, and significantly outperforms state-of-the-art tool-using agents and multi-agent systems built on the same backbone at lower cost.
\end{abstract}

\section{Introduction}
\label{sec:intro}

Tool-augmented medical agents rest on a simple premise: a reasoner that can call expert models---segmenters, detectors, report generators, retrievers---should make better decisions than one that cannot~\cite{li2024mmedagent,fallahpour2025medrax,wang2026medagentpro}. Behind this premise lies an assumption that is rarely stated: anything a tool returns is evidence, and extra evidence is at worst left unused. We find that in medicine this assumption often fails. When all tool outputs are handed to the reasoner, many cases that it originally answered correctly are flipped to wrong (\cref{fig:teaser}a). We call this phenomenon \textbf{evidence hijacking}: once an erroneous tool output enters the context, it overrides the reasoner's otherwise correct judgment. Agents that become worse after gaining tools have also been reported for chest X-rays, medical visual question answering, and drug reasoning~\cite{lou2025cxragent,shen2026evomedagent,jiang2026medvr,xie2025cureagent}, where the failures are usually patched as defects of a particular system. We argue that they share a common locus: not when a tool is called, but after its output enters the reasoner's context.

Three properties of medical tool outputs make hijacking hard to avoid. \emph{First, they fail silently.} When an optic cup--disc segmenter draws a wrong contour on a low-contrast fundus image, it still returns a vertical cup-to-disc ratio whose format and apparent precision are indistinguishable from a correct one; when a report generator hallucinates a consolidation, it does so in the same confident tone as when it does not~\cite{zhang2026radagents,huai2026which}. \emph{Second, they are heterogeneous.} Masks, bounding boxes, free-text reports, and retrieved passages share no common notion of reliability, and no single confidence score can say which of them to trust. \emph{Third, the reasoner cannot look away from them.} Once an output enters the context of a frozen language model, it sways the answer whether or not it should.

Shannon information cannot explain this phenomenon. The output of an imaging tool is a deterministic function of the image, so by the data processing inequality it carries no information about the label beyond what the image already contains~\cite{cover2006elements}. Calling a tool or filtering its output should then neither help nor hurt, yet empirically each can do either. The gap lies in an assumption implicit in Shannon's framework: an observer with unbounded computation, able to extract everything the image contains. A real reasoner is bounded, and what matters is not how much information an input carries but how much of it the reasoner can use. \Vinfo~\cite{xu2020theory} formalizes exactly this quantity, measuring the information about the label that a given predictive family can \emph{use}. Because usability is defined relative to the family, computation can create it: although a segmentation mask is a function of the image, it can make the label easier to read out for a bounded reasoner. This accounts for why tools can help. Why they can also hurt hinges on a precondition of the same theory: usable information is guaranteed to be non-negative only when the predictive family has \emph{optional ignorance}, the ability to disregard any side input. A frozen reasoner lacks exactly this ability, as stated by the third property: whatever enters its context shapes its answer. A misleading tool output can therefore carry \emph{negative} usable information, and evidence hijacking is how this negativity manifests in decisions.

Existing methods do not address this point. Acquisition control decides which tools to call before any output is produced~\cite{ma2026activemedagent,wu2026from}, so it cannot choose by content, and once a call is made, whatever the tool returns enters the reasoner's context in full. Verification and reflection take place inside that context, where hijacking has already happened before any judgment is made. Neither line treats what counts as evidence as a decision in its own right, made apart from the reasoner that answers.

We therefore recast tool-augmented medical reasoning as \textbf{evidence governance}, which makes this decision explicitly and outside the reasoner. This mirrors a court of law, which separates ruling on evidence from deciding the case: the judge rules on admissibility and the jury hears only what has been admitted. Our \textbf{\methodfull} (\textbf{\method}) follows the same division (\cref{fig:framework}). Every tool output, whether a mask, a report, or a retrieved passage, is written as a \textbf{tool card} that can be checked against the input; a \textbf{Checker} that reads only the input and the cards, and never enters the context of the \textbf{Solver}, rules on which cards are admitted; and the Solver answers from the admitted cards alone.

\Vinfo yields two guarantees for this design. First, governance restores optional ignorance outside the reasoner, so the best the governed system can attain is never worse than the better of admitting everything and admitting nothing. Second, the admission gain splits into a budget term, from letting the reasoner see \emph{less}, and a selection term, from letting it see \emph{better}, and only a gate that reads card content can obtain the latter (\cref{sec:theory-cov}). The right reference for evaluating an evidence filter is hence a random gate that admits the same number of cards per case, not admitting all of them.

Our contributions are as follows:
\begin{itemize}[leftmargin=*,itemsep=1pt,topsep=2pt]
  \item \textbf{Phenomenon.} We are the first to reveal and systematically quantify evidence hijacking in tool-augmented medical agents, and attribute it to the frozen reasoner's loss of optional ignorance.
  \item \textbf{Theory.} We develop a \Vinfo theory of evidence governance (\cref{sec:theory}), which shows that governance restores optional ignorance outside the reasoner and establishes the budget-matched random gate as the reference for evaluating evidence filters.
  \item \textbf{Method.} We propose \method (\cref{sec:method}), the first evidence governance framework for medical agents, which rules after acquisition on each individual tool output; it requires no training and plugs into frozen backbones.
  \item \textbf{Results.} Across text, fundus, and chest X-ray modalities, \method achieves the highest accuracy among general and medical multimodal LLMs of comparable scale (\cref{fig:teaser}b), and significantly outperforms state-of-the-art tool-using agents and multi-agent systems on the same backbone with fewer tokens and several-fold faster inference on imaging tasks. The more erroneous the tool outputs, the larger its advantage over admitting all of them, and it consistently stays close to the level unaffected by tools. The Checker's admission significantly beats budget-matched random admission on all three tasks, and the first-order characterization computed from single-card usable information predicts the actual gains of different card subsets.
\end{itemize}

\section{Related Work}
\label{sec:related}

\paragraph{Tool-augmented medical agents.}
Medical agents extend language or vision--language reasoners with expert models: they orchestrate segmenters, classifiers, and report generators through instruction tuning and prompting~\cite{li2024mmedagent,fallahpour2025medrax}, often embedded in structured clinical workflows~\cite{wang2026medagentpro}; they train the reasoner to zoom, ground, and search over images at inference time~\cite{jiang2026medvr,wang2026mira}; or they assign roles to multiple agents and aggregate their opinions~\cite{kim2024mdagents,zhou2025mam}. All these lines measure progress by what an agent can \emph{acquire}, and take returned outputs to be evidence by default.

\paragraph{Judging tool outputs.}
Recent work acknowledges that tool outputs can be wrong, but mostly lets the reasoner judge them in place: by reflecting on their plausibility~\cite{wang2026medagentpro,wang2026mira}, annotating each output with supporting and refuting evidence~\cite{lou2025cxragent}, arbitrating when tools conflict~\cite{zhang2026radagents},  or seeking consensus among multiple agents~\cite{kim2024mdagents,zhou2025mam}. Others learn which tool to trust through reinforcement learning or memory~\cite{huai2026which,shen2026evomedagent}, score evidence by whether it lowers the loss of a reference answer~\cite{lu2026medr2}, or verify evidence against the hypothesis the agent is pursuing~\cite{ge2026clinicalagents}. In the former, judgment happens inside the answering model's own context and is conditioned on its tentative hypothesis, a form of self-verification already shown to be unreliable in medical VQA~\cite{jin2026verification}; adding agents does not escape this, since agents that read the same wrong output are misled alike. In the latter, learned trust can be assigned only to a \emph{tool}, not to \emph{the output at hand}.

\paragraph{Controlling what the reasoner sees.}
Adaptive retrieval and active acquisition decide whether to retrieve or which evidence channel to open based on token confidence~\cite{jiang2023active}, disagreement among sampled answers~\cite{wu2026from}, or a learned controller~\cite{ma2026activemedagent}, while selective prediction lets an agent abstain~\cite{li2025atcxr,raiaan2026evade}. They are motivated by the harm of long or irrelevant context~\cite{shi2023large,liu2024lost,xie2025cureagent}, which ActiveMedAgent calls information overload. Evidence hijacking is different: overload stems from seeing \emph{too much}, hijacking from seeing \emph{the wrong thing}. Even with a fixed number of cards per case, admitting all of them gets worse as more of them are wrong. CRAG~\cite{yan2024corrective} and Self-RAG~\cite{asai2024selfrag} filter retrieved documents one by one, but they target text and judge whether a document is \emph{relevant}, whereas the typical failure of a medical tool output is to be relevant yet \emph{wrong}: an incorrect cup-to-disc ratio is highly relevant to glaucoma screening. None of these lines separates the decision of what counts as evidence from the reasoner that answers, which is the premise of evidence governance.

\paragraph{Usable information.}
\Vinfo~\cite{xu2020theory} defines the information that a computationally bounded predictive family can extract. It has been used to measure dataset difficulty~\cite{ethayarajh2022understanding}, to probe representations conditionally~\cite{hewitt2021conditional}, and to evaluate free-text rationales~\cite{chen2023rev}, always to quantify how informative an input is. We are the first to use it to characterize the relationship between an agent and its tools.

\section{Context-Independent Evidence Governance}
\label{sec:method}

\begin{figure*}[t]
  \centering
  \includegraphics[width=0.74\linewidth]{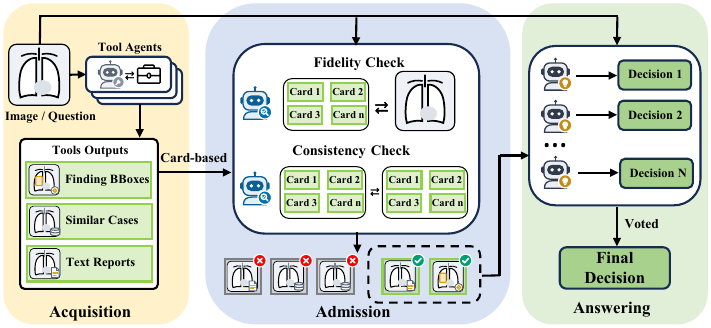}
  \caption{\textbf{\Methodfull (\method).} \emph{Acquisition:} tool agents call their tools on the input and write each result as a verifiable tool card, which shows the finding together with the material it rests on. \emph{Admission:} the Checker checks the cards for fidelity to input and for consistency with one another, and rules on which cards are admitted. \emph{Answering:} the Solver answers from input and the admitted cards alone, and its decisions are combined by vote.}
  \label{fig:framework}
\end{figure*}

\Method (\cref{fig:framework}) realizes evidence governance by splitting tool-augmented reasoning into three steps, acquisition, admission, and answering. Tool agents write their results as verifiable cards, the Checker rules on which cards become evidence outside the Solver's context, and the Solver answers from the admitted cards alone.

\subsection{Problem Setup}
\label{sec:setup}

Each case consists of an input $x$ and a label $y\in\Yc$, where $x$ is a medical image, or a clinical vignette together with a question. A set of external tools, such as segmenters, detectors, report generators, and retrievers, can be applied to $x$. The reasoner is a frozen model $q$. For any evidence set $A$ placed in its context, we write its predictive distribution over $\Yc$ as $q(\cdot\mid x,A)$, and as $q(\cdot\mid x)$ when the context contains no evidence. The task is to decide which evidence may enter $A$.

\subsection{Tool Cards}
\label{sec:cards}

Tool outputs come in mutually incompatible forms: masks, bounding boxes, free-text reports, and retrieved passages. To rule on all of them in one admission procedure, each tool is called by its own tool agent, which writes the result as a card. For an input $x$, this yields the card pool
\begin{equation}
  \Cc=\{c_1,\dots,c_M\}.
  \label{eq:pool}
\end{equation}
Cards are \emph{verifiable}: besides stating a finding, a card displays the material on which the finding rests (a contour overlaid on the original image, a generated report, or a verbatim quoted passage), so that the finding can be compared directly against $x$.

\subsection{Checker and Solver}
\label{sec:checker}

The Checker reads $x$ and the card pool $\Cc$, and returns the set of cards admitted as evidence:
\begin{equation}
  \Cg=g(x,\Cc)\subseteq\Cc.
  \label{eq:gate}
\end{equation}
The Checker is \emph{context-independent}: its ruling depends only on the input and the card pool; it is not conditioned on the Solver's context, hypotheses, or answers, nor does it read the label. Formally, $g$ belongs to the gate family
\begin{equation}
  \Gc=\{\,g:\ \Cg \text{ is determined by } (x,\Cc)\,\},
  \label{eq:family}
\end{equation}
where $g$ may use internal randomness. $\Gc$ contains two trivial gates: $g_0$ admits nothing and $g_1$ admits everything. The Checker examines the pool along two lines: \emph{fidelity}, whether the material shown on a card, set against $x$, supports the finding it states; and \emph{consistency}, whether the cards corroborate or contradict one another. Conflicts among cards mark where to look, while the ruling rests on fidelity to the input, since tools can also agree in error. Both depend on nothing beyond $(x,\Cc)$, which is why $g$ rules on the pool as a whole rather than on one card at a time. The Checker does not know which answer the Solver leans towards, and so has no way to favor cards that agree with a tentative conclusion. The Solver is the reasoner conditioned on the input and the admitted cards:
\begin{equation}
  \hat y\sim q(\cdot\mid x,\Cg).
  \label{eq:solver}
\end{equation}

\section{A \texorpdfstring{$\mathcal{V}$-Information}{V-Information} Theory of Evidence Governance}
\label{sec:theory}

This section explains why a reasoner can be hijacked by its own tools, why governing evidence outside the reasoner corrects this, and how to tell whether a gate selects well or merely makes the reasoner see less. Proofs are deferred to the supplementary material.

\subsection{Losing and Restoring Optional Ignorance}
\label{sec:theory-ignore}

Following the pointwise form of \citet{ethayarajh2022understanding}, the pointwise usable information of an evidence set $A$ for the reasoner $q$ is
\begin{equation}
  v_x(A)=\log q(y\mid x,A)-\log q(y\mid x),
  \label{eq:pvi}
\end{equation}
and the usable information of a gate $g$ is its expectation over cases and gate randomness, $I(g)=\E\big[v_x(\Cg)\big]$. The non-negativity of \Vinfo relies on optional ignorance~\cite{xu2020theory}, which a frozen reasoner does not have: both $v_x(\Cc)$ and $I(g_1)$ can be negative, in which case admitting everything makes it harder for the reasoner to recover the correct label than seeing no card at all. In terms of decisions, this is evidence hijacking. Governance moves optional ignorance outside the reasoner: the object of study is no longer the reasoner $q$ alone but the governed predictive family $\mathcal V_{\Gc}=\{q(\cdot\mid x,\Cg):g\in\Gc\}$, which contains both $g_0$ and $g_1$, so its usable information $\sup_{g\in\Gc}I(g)$ is automatically no lower than either $0$ or $I(g_1)$. Whether a practical gate exceeds both trivial gates is an empirical question; \cref{sec:exp-theory} shows that the Checker does on all three tasks.

\subsection{Budget and Selection}
\label{sec:theory-budget}

Relative to admitting everything, the gain of a gate, $T(g)=I(g)-I(g_1)$, mixes \emph{how many} cards are admitted with \emph{which} ones. Let $\pi(g)$ be the budget-matched random gate of $g$, which on every input draws $|\Cg|$ cards uniformly from the pool. Inserting $I(\pi(g))$ gives
\begin{equation}
  \underbrace{I(g)-I(g_1)}_{T(g)}
  =\underbrace{I(\pi(g))-I(g_1)}_{B(g)}
  +\underbrace{I(g)-I(\pi(g))}_{S(g)}.
  \label{eq:decomp}
\end{equation}
The \emph{budget term} $B$ is the gain from seeing less, and its sign can vary with the task: random pruning helps when the pool overloads the reasoner~\cite{shi2023large}, and hurts when most cards are useful. The \emph{selection term} $S$ compares which cards are admitted under the same budget, and only $S>0$ shows that a gate has identified better evidence. If a gate looks at neither the content nor the source of cards and depends only on $x$, the pool size, card positions, and internal randomness---as with randomly dropping cards, truncating the context, or deciding whether to add evidence from the reasoner's own confidence---we call it a \emph{content-agnostic gate}. When the pool is randomly ordered, the set that a content-agnostic gate admits for a given count is already a uniform draw, distributed identically to $\pi(g)$; hence $S=0$. The gain of an evidence filter over admitting everything may therefore come entirely from seeing less; to show that it selects well, the reference must be $\pi(g)$, not $g_1$.

\subsection{Where the Selection Gain Comes From}
\label{sec:theory-cov}

Applying \cref{eq:pvi} to a single card, $v_x(c)=v_x(\{c\})$ is the usable information of card $c$ presented alone; it is positive if $c$ helps the reasoner and negative if $c$ misleads it. For an input $x$, let $K=|\Cc|$ and $a_c=\mathbf{1}[c\in\Cg]$.

\begin{proposition}[First-order characterization of the selection gain]
\label{prop:cov}
\begin{equation}
  S(g)\ \approx\ \E\Big[\,K\cdot\operatorname{Cov}_{c\in\Cc}\big(a_c,\ v_x(c)\big)\Big],
  \label{eq:cov}
\end{equation}
where the covariance is taken uniformly over the cards in the pool. Equality holds exactly when the gate admits at most one card or all cards per case; in general, the two sides differ only by interactions among admitted cards.
\end{proposition}

\Cref{eq:cov} shows that, under the same budget, $S>0$ when admission decisions correlate positively with single-card usable information. A gate need not estimate the value of $v_x(c)$; it only needs to rank cards that help the reasoner above cards that mislead it. A content-agnostic gate admits cards independently of what they say, so the covariance vanishes, consistent with \cref{sec:theory-budget}. \Cref{eq:cov} also makes the theory testable: once single-card usable information is measured, the selection gain of a gate can be predicted before the gate is run. We test this prediction in \cref{sec:exp-theory}.

\section{Experiments}
\label{sec:exp}

We evaluate \method on three modalities. We first compare it with existing medical models and agent systems, then test its robustness to evidence hijacking, examine whether the theory of \cref{sec:theory} accounts for its gains, and finally ablate each design choice.

\begin{table*}[t]
  \centering
  \caption{\textbf{Main results.} bAcc on imaging tasks; accuracy (\%) on the three medical QA datasets. \sig: \method is significantly better ($p<0.05$). Best per column in bold. General and medical MLLMs, like ours, take the majority vote over 8 samples; the other baselines follow their original pipelines. $^\ast$Training data include chest X-rays and may overlap in source with the test set. $^\S$Official 13B weights. $^\ddagger$Reproduced with official settings on the text-only Qwen3-8B~\cite{yang2025qwen3}, as Qwen3-VL-8B lacks the thinking mode its retrieval queries require.}
  \label{tab:main}
  \small
  \setlength{\tabcolsep}{9pt}
  \setlength{\aboverulesep}{0pt}
  \setlength{\belowrulesep}{0pt}
  \renewcommand{\arraystretch}{1.0}
  \newcommand{\grp}[1]{\rowcolor{grpgray}\multicolumn{6}{@{}l}{\hspace{3pt}\textit{\color{black!65}#1}}\\}
  \begin{tabular}{@{}l cc ccc@{}}
    \toprule
    & \multicolumn{2}{c}{\cellcolor{imgteal!12}\textcolor{imgteal}{\textbf{\textit{Imaging}\,$\cdot$\,bAcc}}}
    & \multicolumn{3}{c}{\cellcolor{oursblue!10}\textcolor{oursblue}{\textbf{\textit{Text}\,$\cdot$\,Acc (\%)}}} \\
    \arrayrulecolor{imgteal}\cmidrule(lr){2-3}\arrayrulecolor{oursblue}\cmidrule(l){4-6}\arrayrulecolor{black}
    Method & Glaucoma & Chest X-ray & MedQA & MedMCQA & NEJM \\
    \midrule
    \grp{General MLLMs}
    \hspace{3pt}Qwen3-VL-4B            & 0.620\sig & 0.776\sig & 66.77\sig & 58.71\sig & 47.94\sig \\
    \hspace{3pt}Qwen3-VL-8B (backbone) & 0.636\sig     & 0.676\sig & 74.94\sig & 64.40\sig & 55.73\sig \\
    \hspace{3pt}InternVL3-8B           & 0.511\sig & 0.692\sig & 64.34\sig & 56.61\sig & 46.87\sig \\
    \grp{Medical MLLMs}
    \hspace{3pt}HuatuoGPT-Vision-7B    & 0.509\sig & 0.634\sig & 42.66\sig & 44.80\sig & 36.18\sig \\
    \hspace{3pt}Lingshu-7B$^\ast$      & 0.545\sig & 0.798     & 60.49\sig & 48.34\sig & 38.78\sig \\
    \hspace{3pt}MedGemma-4B            & 0.500\sig & 0.502\sig & 64.34\sig & 55.15\sig & 47.18\sig \\
    \hspace{3pt}Hulu-Med-14B$^\ast$    & 0.611\sig & 0.822     & 77.06\sig & 60.94\sig & 50.08\sig \\
    \grp{Task-specific methods}
    \hspace{3pt}BiomedCLIP (zero-shot) & 0.464\sig & 0.653\sig & --        & --        & --        \\
    \hspace{3pt}CheXagent              & --        & 0.759\sig & --        & --        & --        \\
    \hspace{3pt}RAG                    & --        & --        & 76.20\sig & 68.75\sig & 61.68     \\
    \hspace{3pt}CRAG                   & --        & --        & 73.61\sig & 62.75\sig & 53.59\sig \\
    \hspace{3pt}Self-RAG$^\S$          & --        & --        & 43.44\sig & 44.47\sig & 39.08\sig \\
    \hspace{3pt}MA-RAG$^\ddagger$      & --        & --        & 76.67\sig & 67.34\sig & 59.85\sig \\
    \grp{Agents on the same backbone}
    \hspace{3pt}ReAct                  & 0.544\sig & 0.754\sig & 77.45\sig & 65.03\sig & 59.39\sig \\
    \hspace{3pt}MedAgents              & 0.650\sig & 0.720\sig & 78.00\sig & 65.86\sig & 59.85\sig \\
    \hspace{3pt}MDAgents               & 0.586\sig & 0.700\sig & 78.00\sig & 65.89\sig & 61.22\sig \\
    \hspace{3pt}MAM                    & 0.627\sig & 0.760\sig & 77.22\sig & 68.73\sig & 61.98     \\
    \midrule
    \rowcolor{oursblue!10}
    \hspace{3pt}\textcolor{oursblue}{\textbf{Ours}}
      & \textcolor{oursblue}{\best{0.730}} & \textcolor{oursblue}{\best{0.837}}
      & \textcolor{oursblue}{\best{82.72}} & \textcolor{oursblue}{\best{70.67}} & \textcolor{oursblue}{\best{65.19}} \\
    \bottomrule
  \end{tabular}
\end{table*}

\subsection{Experimental Setup}
\label{sec:exp-setup}

\paragraph{Tasks and data.}
The three tasks cover three kinds of tool output: masks, reports, and retrieved passages. \emph{Fundus glaucoma screening} uses the 400 fundus images of the REFUGE2 validation set~\cite{fang2022refuge2}, with optic disc--cup segmentations as tool outputs. \emph{Chest X-ray pneumonia classification} uses the official test set of the Kermany pediatric chest X-ray dataset~\cite{kermany2018identifying}, deduplicated by patient to 431 images, with structured findings, free-text reports, and similar-case retrieval as tool outputs. \emph{Medical question answering} uses MedQA~\cite{jin2021what} (1,273 questions), MedMCQA~\cite{pal2022medmcqa} (4,183), and NEJM~\cite{katz2024gpt} (655), with retrieved passages as tool outputs. We report balanced accuracy (bAcc) on the imaging tasks. On question answering, \cref{tab:main} reports accuracy on each dataset, and the other tables report the macro-average over the three datasets, weighted equally.

\paragraph{Tool pools.}
For glaucoma, BiomedParse~\cite{zhao2025foundation}, MedSAM~\cite{ma2024segment}, and SAM~\cite{kirillov2023segment} each produce an optic disc--cup segmentation; a card shows the contour overlaid on the original image and the resulting vertical cup-to-disc ratio (vCDR). For chest X-rays, ChEX~\cite{muller2024chex}, MAIRA-2~\cite{bannur2024maira2}, CURE~\cite{messina2026cure}, and similar-case retrieval based on RAD-DINO~\cite{perezgarcia2025exploring} each provide a set of findings or a report. For question answering, a tool agent writes retrieval queries from the question; passages are recalled by BM25~\cite{robertson2009probabilistic} from the MedCorp corpus (PubMed, StatPearls, textbooks, and Wikipedia)~\cite{xiong2024benchmarking} and reranked by MedCPT~\cite{jin2023medcpt}, and the top 6 per query become candidate cards, each quoting its passage verbatim.

\paragraph{Implementation.}
All components are frozen and nothing is trained. Both the Checker and the Solver are Qwen3-VL-8B~\cite{bai2025qwen3vl}; the Solver draws 8 samples and takes the majority vote~\cite{wang2023selfconsistency}. The same 8-sample vote is applied to every admission policy and Solver variant we compare and to all general and medical MLLM baselines; agent systems and retrieval-augmented methods are run with their original pipelines, so that each follows its published protocol. Prompts, card formats, and the Checker's output format for each task are given in the supplementary material.

\paragraph{Baselines.}
We compare with four groups of methods. (i)~\emph{General multimodal LLMs}: Qwen3-VL-4B, Qwen3-VL-8B (our backbone)~\cite{bai2025qwen3vl}, and InternVL3-8B~\cite{zhu2025internvl3}. (ii)~\emph{Medical multimodal LLMs}: HuatuoGPT-Vision-7B~\cite{chen2024towards}, Lingshu-7B~\cite{xu2026lingshu}, MedGemma-4B~\cite{sellergren2026open}, and Hulu-Med-14B~\cite{jiang2025hulumed}. (iii)~\emph{Task-specific methods}: zero-shot BiomedCLIP~\cite{zhang2025multimodal} and the chest X-ray model CheXagent~\cite{chen2024chexagent} for imaging; and, for question answering, retrieval-augmented methods on the same corpus: RAG~\cite{lewis2020retrieval}, CRAG~\cite{yan2024corrective}, Self-RAG~\cite{asai2024selfrag}, and MA-RAG~\cite{wu2026from}. (iv)~\emph{Agent systems on the same backbone}: a ReAct agent~\cite{yao2023react} over our tool pool (in the style of MedRAX~\cite{fallahpour2025medrax}), and the multi-agent systems MedAgents~\cite{tang2024medagents}, MDAgents~\cite{kim2024mdagents}, and MAM~\cite{zhou2025mam}. All agent baselines use Qwen3-VL-8B as the backbone and read the same tool cards as we do, only without admission.

\paragraph{Statistics.}
All comparisons are paired and use a stratified bootstrap~\cite{efron1993introduction} with 5,000 resamples, stratified by class on imaging tasks and by dataset on the question-answering macro-average; per-dataset results are tested within each dataset. 

\subsection{Main Results}
\label{sec:exp-main}

\paragraph{\Method attains the highest accuracy on all three tasks.}
On the same backbone, it outperforms the tool-using agents and multi-agent systems in 19 of the 20 per-dataset comparisons with significance, with relative gains of up to 34.2\% over ReAct and 16.4\% over MAM, the strongest multi-agent system. Since these systems read exactly the same tool cards, the improvement comes not from more or better tools but from how the same evidence is handled; \cref{sec:exp-theory} isolates the share due to admission. \method also surpasses every multimodal LLM, including Hulu-Med-14B with nearly twice the parameters, and all four RAG variants.

\begin{table}[t]
  \centering
  \caption{\textbf{Per-case cost.} Tokens: prompt plus generation per case over the full test set. Time: median end-to-end latency (s) on a 50-case subset.}
  \label{tab:cost}
  \small
  \setlength{\tabcolsep}{3.2pt}
  \begin{tabular}{@{}lcccccc@{}}
    \toprule
    & \multicolumn{2}{c}{Glaucoma} & \multicolumn{2}{c}{Chest X-ray} & \multicolumn{2}{c}{Med.\ QA} \\
    \cmidrule(lr){2-3}\cmidrule(lr){4-5}\cmidrule(l){6-7}
    Method & Tokens & Time & Tokens & Time & Tokens & Time \\
    \midrule
    MedAgents & 55.9k  & 39.6  & 82.8k & 50.2 & 72.8k & \best{37.9} \\
    MDAgents  & 164.6k & 120.4 & 40.8k & 19.7 & 78.3k & 155.1 \\
    MAM       & 36.6k  & 51.0  & 54.9k & 52.7 & 38.6k & 50.2 \\
    \textbf{Ours} & \best{25.3k} & \best{5.5} & \best{33.3k} & \best{8.0} & \best{34.6k} & 39.0 \\
    \bottomrule
  \end{tabular}
\end{table}

\paragraph{More accurate, and also cheaper.}
On all three tasks, \method consumes fewer tokens per case than every multi-agent system, using only about one-sixth of MDAgents' budget on glaucoma (\cref{tab:cost}). Multi-agent systems have several roles read and discuss all tool outputs repeatedly, whereas our Checker reads the cards once and the Solver's 8 samples are independent and can run in parallel. As a result, its end-to-end latency on imaging tasks is 2.5--22$\times$ lower than that of the multi-agent systems.

\subsection{Admission Withstands Evidence Hijacking}
\label{sec:exp-hijack}

\begin{figure}[t]
  \centering
  \includegraphics[width=0.88\linewidth]{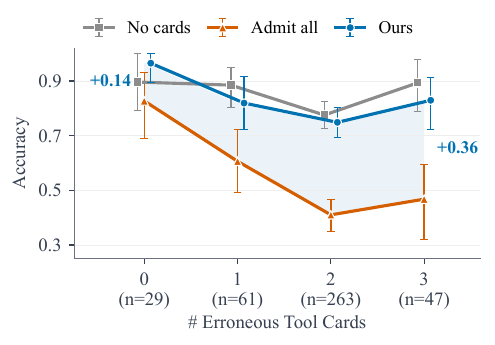}
  \caption{\textbf{Hijacking grows with erroneous evidence.} Glaucoma screening accuracy on 400 cases, grouped by the number of erroneous tool cards per case (out of 3). Gray: no cards; orange: admit all; blue: \method. Error bars are 95\% bootstrap confidence intervals; the shaded band marks the gap between admit-all and ours.}
  \label{fig:hijack}
\end{figure}

\paragraph{The cost of admitting all tool outputs grows quickly with erroneous outputs; the advantage of \method grows with it, while staying close to the level unaffected by tools.}
In glaucoma screening, handing all three segmentation cards to the Solver flips 178 originally correct cases (44.5\%) to wrong. To see how hijacking grows with erroneous evidence, we stratify cases by the number of erroneous cards, where a card is erroneous if its vCDR differs from the ground truth by more than 0.10. With the same card pool and Solver, we compare three admission policies: no cards, admit all, and \method (\cref{fig:hijack}).

Admit-all collapses from 0.83 to 0.61 with a single erroneous card and to below 0.5 with more, so the advantage of \method widens from $+$0.14 to $+$0.36, by 0.086 per erroneous card ($p=0.022$, robust to controlling for class). Meanwhile, \method stays within 0.07 of no cards at every level, with no trend in this gap ($p=0.29$), and exceeds both baselines when all cards are correct. This matches \cref{sec:theory-ignore}: admission restores optional ignorance outside the reasoner, so the system falls back to the undisturbed level when cards are wrong and exploits them when they are reliable.

Because every case has exactly three cards, the horizontal axis of \cref{fig:hijack} varies only the fraction of erroneous cards, not context length: admit-all fails from seeing \emph{the wrong thing}, not \emph{too much}, which separates hijacking from information overload~\cite{ma2026activemedagent} and cannot be repaired by limiting how much the reasoner sees.

\subsection{Theory Meets Practice}
\label{sec:exp-theory}

\Cref{sec:theory} makes two testable predictions: the gain of admission over admitting everything splits into a budget term and a selection term, of which only content-based gates obtain the latter (\cref{eq:decomp}); and the selection term is determined by the covariance between admission decisions and single-card usable information (\cref{eq:cov}). We test them in turn.

\begin{table}[t]
  \centering
  \caption{\textbf{Same card pool, same Solver; only the admitted cards change.} Metrics as in \cref{tab:main}. \emph{No cards} uses the same pipeline and answer template as ours, with the Checker rejecting every card. \emph{Budget-matched random} admits, per case, as many random cards as ours. \emph{Rejected only} admits only the cards the Checker rejects.}
  \label{tab:gates}
  \small
  \setlength{\tabcolsep}{3.5pt}
  \begin{tabular}{@{}lccc@{}}
    \toprule
    Admission & Glaucoma & Chest X-ray & Med.\ QA \\
    \midrule
    No cards $g_0$                  & 0.636\sig & 0.676\sig & 65.02\sig \\
    Admit all $g_1$                 & 0.627\sig & 0.821\sig     & 72.31\sig \\
    Budget-matched rand.\ $\pi(g)$  & 0.637\sig & 0.803\sig & 71.82\sig \\
    Rejected only                   & 0.659\sig & 0.579\sig & 69.66\sig \\
    \midrule
    \textbf{Ours}                   & \best{0.730} & \best{0.837} & \best{72.86} \\
    \bottomrule
  \end{tabular}
\end{table}

\begin{figure*}[t]
  \centering
  \makebox[0.379\linewidth][l]{\small\textbf{(a)}}\hspace{0.03\linewidth}\makebox[0.481\linewidth][l]{\small\textbf{(b)}}\par\vspace{1pt}
  \includegraphics[width=0.379\linewidth]{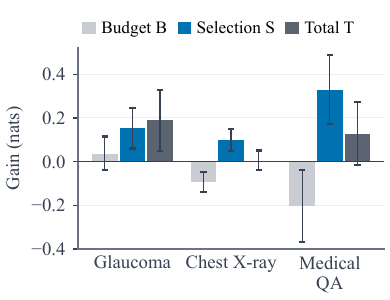}\hspace{0.03\linewidth}%
  \includegraphics[width=0.481\linewidth]{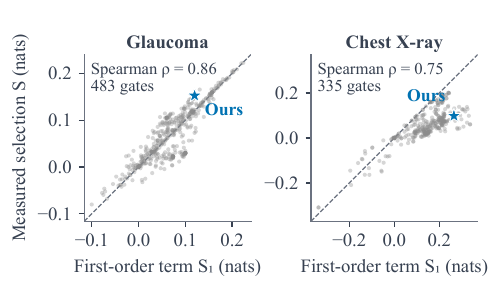}
  \caption{\textbf{Theory meets practice.} (a)~Budget term $B$, selection term $S$, and total gain $T$ of \cref{eq:decomp} on the three tasks (nats, 95\% CI). On imaging tasks, all card subsets of each case are enumerated to compute $I(\pi(g))$ exactly; on question answering, it is estimated by budget-matched random sampling with a single Solver sample. (b)~First-order term of \cref{eq:cov} versus the measured selection term. Each point is a gate (483 on glaucoma, 335 on chest X-ray); the star marks \method; the dashed line is $y=x$.}
  \label{fig:theory}
\end{figure*}

\Cref{sec:theory-ignore} guarantees that the best gate in $\Gc$ is never worse than admitting nothing or everything, and the Checker realizes this in practice: \method significantly exceeds both $g_0$ and $g_1$ on all three tasks (\cref{tab:gates}). Which trivial gate is better depends on the task---no cards on glaucoma, admit all on chest X-rays and QA---so any fixed choice loses heavily on some task, whereas \method needs no such choice. Its margin over the better trivial gate is largest where the tools err most often: 0.094 on glaucoma, against 0.016 on chest X-rays and 0.55 points on QA, where most cards are useful (\cref{fig:theory}a). \Method also exceeds budget-matched random admission on all three tasks. Since both admit the same number of cards per case, the gain comes from \emph{which} cards are admitted, not from seeing less. Conversely, admitting only the rejected cards is significantly worse than \method on all three tasks, and on chest X-rays even falls 0.096 below no cards.

\paragraph{The budget--selection decomposition explains why the total gains differ across tasks.}
\Cref{fig:theory}a reports the three terms of \cref{eq:decomp} in usable information (\cref{eq:pvi}, nats). The selection term $S$ is significantly positive on all three tasks. The budget term $B$, in contrast, changes sign across tasks. On glaucoma it is close to zero, and almost all of the selection gain turns into total gain over admit-all. On chest X-rays and QA most cards are useful, so seeing less has a cost that offsets most of the selection gain; this is why the accuracy of \method is close to that of admit-all there (\cref{tab:gates}). Matching accuracy thus does not mean that the Checker does nothing: it selects the right cards while seeing fewer, and the two effects cancel. 

\paragraph{Single-card usable information predicts the selection gain of a gate.}
To test \cref{eq:cov}, we construct a family of gates by sweeping thresholds on the scores of several Checkers, injecting noise of varying strength into the scores, and, on chest X-rays, sweeping separate thresholds for the two stances. For each gate we compute the right-hand side of \cref{eq:cov} from single-card usable information alone, and compare it with the measured selection term (\cref{fig:theory}b). Their rank correlation is 0.86 on glaucoma (483 gates) and 0.75 on chest X-rays (335 gates), and 97\% of the chest X-ray gates agree in sign. On glaucoma, \method admits exactly one card, all cards, or no card on 71\% of cases, where \cref{eq:cov} holds exactly. The selection gain is therefore determined by an interpretable quantity that can be measured before the gate is run: whether admission decisions correlate positively with single-card usable information.

\subsection{Ablation: Every Design Choice Matters}
\label{sec:exp-ablation}

\begin{table}[t]
  \centering
  \caption{\textbf{Where to judge tool outputs.} Same backbone and tool pool.}
  \label{tab:where}
  \small
  \setlength{\tabcolsep}{4pt}
  \begin{tabular}{@{}lccc@{}}
    \toprule
    Configuration & Glaucoma & Chest X-ray & Med.\ QA \\
    \midrule
    No cards $g_0$                 & 0.636\sig & 0.676\sig & 65.02\sig \\
    ReAct (judge in context)             & 0.544\sig & 0.754\sig & 67.29\sig \\
    Solver calls tools itself            & 0.606\sig & 0.814     & 69.11\sig \\
    \midrule
    \textbf{Ours}                        & \best{0.730} & \best{0.837} & \best{72.86} \\
    \bottomrule
  \end{tabular}
\end{table}

\paragraph{Judging tool outputs inside the answering context is unreliable; moving judgment outside the reasoner is key.}
ReAct lets the same model call tools, read their outputs, and answer, all within its own context. On glaucoma it is even 0.092 below no cards: once a wrong segmentation enters the context, the reasoner can no longer ignore it. \Method significantly outperforms ReAct on all three tasks. Letting the Solver call tools itself and read the raw outputs improves on ReAct but remains below \method.

\paragraph{The Checker identifies erroneous cards, and verifiable cards make it more accurate.}
On glaucoma the quality of every card is known (\cref{tab:checker}). Within each class, the Checker's admission score separates correct from erroneous segmentations with AUROC 0.93 and 0.85, admitting most correct cards but only about one in five erroneous ones. Comparing against the original image matters: given only the cards, the AUROC falls significantly. Even then, the Checker still obtains a positive selection term, so the gain identified in \cref{sec:theory} comes from judging content.

\begin{table}[t]
  \centering
  \caption{\textbf{Checker quality on glaucoma.} A card is erroneous if its vCDR deviates from the ground truth by more than 0.10. AUROC and admission rates (Adm., \%) of correct and erroneous cards are within class (glaucomatous/non-glaucomatous). $S$ is the selection term of \cref{eq:decomp}.}
  \label{tab:checker}
  \small
  \setlength{\tabcolsep}{1.6pt}
  \begin{tabular}{@{}lcccc@{}}
    \toprule
    Checker input & AUROC$\uparrow$ & Adm.\ corr.$\uparrow$ & Adm.\ err.$\downarrow$ & $S\uparrow$ \\
    \midrule
    Random gate           & 0.50/0.50 & --    & --    & 0 \\
    Cards only            & 0.87\sig/0.81\sig & 83/74 & 27/31 & $+$0.105\sig \\
    Input$+$cards (ours) & \best{0.93}/\best{0.85} & \best{87}/\best{76} & \best{19}/\best{22} & \best{$+$0.152} \\
    \bottomrule
  \end{tabular}
\end{table}

\section{Conclusion}
\label{sec:conclusion}

Tool outputs are not evidence by default; what counts as evidence should be governed, not left to the reasoner that answers. We showed that medical agents suffer from evidence hijacking, traced it to a frozen reasoner's loss of optional ignorance, and restored that option outside the reasoner with \method, an evidence governance framework of verifiable tool cards, a context-independent Checker, and a Solver that answers only from admitted cards, all without training. \VInfo further separates seeing less from seeing better, which makes the budget-matched random gate the proper reference for any evidence filter. Across three modalities, \method yields the best accuracy at comparable scale and lower cost than agents that read the same tool outputs, and its admission significantly exceeds both admitting every tool output and admitting none on every task.

\section*{LLM Usage Statement}
\label{sec:llm-usage}

We used Claude (Anthropic), an LLM assistant, in three parts of this work. \emph{Writing:} the authors drafted the paper, and the assistant was used to translate the draft into English and to edit it for grammar, wording, and clarity; all scientific content, claims, and reported numbers were provided and checked by the authors. \emph{Code:} the assistant helped write and debug experiment and plotting scripts; the authors designed all experiments, reviewed the code, and verified every reported result. \emph{Related work:} the assistant helped search for and organize related literature and check citation metadata; every cited work was verified by the authors against its original source. LLMs were not used to generate research ideas, theoretical results, or data. Separately, Qwen3-VL-8B and the other models listed in \cref{sec:exp-setup} are components or baselines of the evaluated systems; they are part of the experiments, not writing aids.

{
    \small
    \bibliographystyle{ieeenat_fullname}
    \bibliography{references}

\begin{thebibliography}{58}
\providecommand{\natexlab}[1]{#1}
\providecommand{\url}[1]{\texttt{#1}}
\expandafter\ifx\csname urlstyle\endcsname\relax
  \providecommand{\doi}[1]{doi: #1}\else
  \providecommand{\doi}{doi: \begingroup \urlstyle{rm}\Url}\fi

\bibitem[Asai et~al.(2024)Asai, Wu, Wang, Sil, and Hajishirzi]{asai2024selfrag}
Akari Asai, Zeqiu Wu, Yizhong Wang, Avirup Sil, and Hannaneh Hajishirzi.
\newblock Self-{RAG}: Learning to retrieve, generate, and critique through
  self-reflection.
\newblock In \emph{The Twelfth International Conference on Learning
  Representations (ICLR)}, 2024.

\bibitem[Bai et~al.(2025)Bai, Cai, Chen, Chen, Chen, Cheng, Deng, Ding, Gao,
  Ge, et~al.]{bai2025qwen3vl}
Shuai Bai, Yuxuan Cai, Ruizhe Chen, Keqin Chen, Xionghui Chen, Zesen Cheng,
  Lianghao Deng, Wei Ding, Chang Gao, Chunjiang Ge, et~al.
\newblock {Qwen3-VL} technical report.
\newblock \emph{arXiv preprint arXiv:2511.21631}, 2025.

\bibitem[Bannur et~al.(2024)Bannur, Bouzid, Castro, Schwaighofer, Thieme,
  Bond-Taylor, Ilse, P{\'e}rez-Garc{\'i}a, Salvatelli, Sharma,
  et~al.]{bannur2024maira2}
Shruthi Bannur, Kenza Bouzid, Daniel~C. Castro, Anton Schwaighofer, Anja
  Thieme, Sam Bond-Taylor, Maximilian Ilse, Fernando P{\'e}rez-Garc{\'i}a,
  Valentina Salvatelli, Harshita Sharma, et~al.
\newblock {MAIRA-2}: Grounded radiology report generation.
\newblock \emph{arXiv preprint arXiv:2406.04449}, 2024.

\bibitem[Chen et~al.(2023)Chen, Brahman, Ren, Ji, Choi, and
  Swayamdipta]{chen2023rev}
Hanjie Chen, Faeze Brahman, Xiang Ren, Yangfeng Ji, Yejin Choi, and Swabha
  Swayamdipta.
\newblock {REV}: Information-theoretic evaluation of free-text rationales.
\newblock In \emph{Proceedings of the 61st Annual Meeting of the Association
  for Computational Linguistics (Volume 1: Long Papers)}, pages 2007--2030,
  Toronto, Canada, 2023. Association for Computational Linguistics.

\bibitem[Chen et~al.(2024{\natexlab{a}})Chen, Gui, Ouyang, Gao, Chen, Chen,
  Wang, Cai, Ji, Wan, and Wang]{chen2024towards}
Junying Chen, Chi Gui, Ruyi Ouyang, Anningzhe Gao, Shunian Chen, Guiming~Hardy
  Chen, Xidong Wang, Zhenyang Cai, Ke Ji, Xiang Wan, and Benyou Wang.
\newblock Towards injecting medical visual knowledge into multimodal {LLMs} at
  scale.
\newblock In \emph{Proceedings of the 2024 Conference on Empirical Methods in
  Natural Language Processing (EMNLP)}, pages 7346--7370, Miami, Florida, USA,
  2024{\natexlab{a}}. Association for Computational Linguistics.

\bibitem[Chen et~al.(2024{\natexlab{b}})Chen, Varma, Delbrouck, Paschali,
  Blankemeier, Van~Veen, Valanarasu, Youssef, Cohen, Reis,
  et~al.]{chen2024chexagent}
Zhihong Chen, Maya Varma, Jean-Benoit Delbrouck, Magdalini Paschali, Louis
  Blankemeier, Dave Van~Veen, Jeya Maria~Jose Valanarasu, Alaa Youssef,
  Joseph~Paul Cohen, Eduardo~Pontes Reis, et~al.
\newblock {CheXagent}: Towards a foundation model for chest {X}-ray
  interpretation.
\newblock \emph{arXiv preprint arXiv:2401.12208}, 2024{\natexlab{b}}.

\bibitem[Cover and Thomas(2006)]{cover2006elements}
Thomas~M. Cover and Joy~A. Thomas.
\newblock \emph{Elements of Information Theory}.
\newblock Wiley-Interscience, Hoboken, NJ, 2nd edition, 2006.

\bibitem[Efron and Tibshirani(1993)]{efron1993introduction}
Bradley Efron and Robert~J. Tibshirani.
\newblock \emph{An Introduction to the Bootstrap}.
\newblock Chapman \& Hall, New York, 1993.

\bibitem[Ethayarajh et~al.(2022)Ethayarajh, Choi, and
  Swayamdipta]{ethayarajh2022understanding}
Kawin Ethayarajh, Yejin Choi, and Swabha Swayamdipta.
\newblock Understanding dataset difficulty with $\mathcal{V}$-usable
  information.
\newblock In \emph{Proceedings of the 39th International Conference on Machine
  Learning (ICML)}, pages 5988--6008. PMLR, 2022.

\bibitem[Fallahpour et~al.(2025)Fallahpour, Ma, Munim, Lyu, and
  Wang]{fallahpour2025medrax}
Adibvafa Fallahpour, Jun Ma, Alif Munim, Hongwei Lyu, and Bo Wang.
\newblock {MedRAX}: Medical reasoning agent for chest x-ray.
\newblock In \emph{Proceedings of the 42nd International Conference on Machine
  Learning}, pages 15661--15676. PMLR, 2025.

\bibitem[Fang et~al.(2022)Fang, Li, Wu, Fu, Sun, Son, Yu, Zhang, Yuan, Bian,
  et~al.]{fang2022refuge2}
Huihui Fang, Fei Li, Junde Wu, Huazhu Fu, Xu Sun, Jaemin Son, Shuang Yu, Menglu
  Zhang, Chenglang Yuan, Cheng Bian, et~al.
\newblock {REFUGE2} challenge: A treasure trove for multi-dimension analysis
  and evaluation in glaucoma screening.
\newblock \emph{arXiv preprint arXiv:2202.08994}, 2022.

\bibitem[Ge et~al.(2026)Ge, Li, Wang, Hu, Zhang, and Li]{ge2026clinicalagents}
Zhuohan Ge, Haoyang Li, Yubo Wang, Nicole Hu, Chen~Jason Zhang, and Qing Li.
\newblock {ClinicalAgents}: Multi-agent orchestration for clinical decision
  making with dual-memory.
\newblock In \emph{Proceedings of the 32nd ACM SIGKDD Conference on Knowledge
  Discovery and Data Mining V.2}, pages 10969--10980. ACM, 2026.

\bibitem[Hewitt et~al.(2021)Hewitt, Ethayarajh, Liang, and
  Manning]{hewitt2021conditional}
John Hewitt, Kawin Ethayarajh, Percy Liang, and Christopher Manning.
\newblock Conditional probing: measuring usable information beyond a baseline.
\newblock In \emph{Proceedings of the 2021 Conference on Empirical Methods in
  Natural Language Processing (EMNLP)}, pages 1626--1639, Online and Punta
  Cana, Dominican Republic, 2021. Association for Computational Linguistics.

\bibitem[Huai et~al.(2026)Huai, Yang, and Li]{huai2026which}
Zheang Huai, Honglong Yang, and Xiaomeng Li.
\newblock Which tool response should {I} trust? {Tool-Expertise-Aware} chest
  {X}-ray agent with multimodal agentic learning.
\newblock \emph{arXiv preprint arXiv:2602.21517}, 2026.

\bibitem[Jiang et~al.(2025)Jiang, Wang, Song, Hu, Zhou, Pu, Zhang, Yang, Feng,
  Zhou, et~al.]{jiang2025hulumed}
Songtao Jiang, Yuan Wang, Sibo Song, Tianxiang Hu, Chenyi Zhou, Bin Pu, Yan
  Zhang, Zhibo Yang, Yang Feng, Joey~Tianyi Zhou, et~al.
\newblock {Hulu-Med}: A transparent generalist model towards holistic medical
  vision-language understanding.
\newblock \emph{arXiv preprint arXiv:2510.08668}, 2025.

\bibitem[Jiang et~al.(2023)Jiang, Xu, Gao, Sun, Liu, Dwivedi-Yu, Yang, Callan,
  and Neubig]{jiang2023active}
Zhengbao Jiang, Frank Xu, Luyu Gao, Zhiqing Sun, Qian Liu, Jane Dwivedi-Yu,
  Yiming Yang, Jamie Callan, and Graham Neubig.
\newblock Active retrieval augmented generation.
\newblock In \emph{Proceedings of the 2023 Conference on Empirical Methods in
  Natural Language Processing (EMNLP)}, pages 7969--7992, Singapore, 2023.
  Association for Computational Linguistics.

\bibitem[Jiang et~al.(2026)Jiang, Guo, Fang, Xiao, Hu, Sun, and
  Xu]{jiang2026medvr}
Zheng Jiang, Heng Guo, Chengyu Fang, Changchen Xiao, Xinyang Hu, Lifeng Sun,
  and Minfeng Xu.
\newblock {MedVR}: Annotation-free medical visual reasoning via agentic
  reinforcement learning.
\newblock In \emph{The Fourteenth International Conference on Learning
  Representations}, 2026.

\bibitem[Jin et~al.(2021)Jin, Pan, Oufattole, Weng, Fang, and
  Szolovits]{jin2021what}
Di Jin, Eileen Pan, Nassim Oufattole, Wei-Hung Weng, Hanyi Fang, and Peter
  Szolovits.
\newblock What disease does this patient have? {A} large-scale open domain
  question answering dataset from medical exams.
\newblock \emph{Applied Sciences}, 11\penalty0 (14):\penalty0 6421, 2021.

\bibitem[Jin et~al.(2023)Jin, Kim, Chen, Comeau, Yeganova, Wilbur, and
  Lu]{jin2023medcpt}
Qiao Jin, Won Kim, Qingyu Chen, Donald~C. Comeau, Lana Yeganova, W.~John
  Wilbur, and Zhiyong Lu.
\newblock {MedCPT}: Contrastive pre-trained transformers with large-scale
  {PubMed} search logs for zero-shot biomedical information retrieval.
\newblock \emph{Bioinformatics}, 39\penalty0 (11):\penalty0 btad651, 2023.

\bibitem[Jin et~al.(2026)Jin, Zhao, Kang, Zhang, and Li]{jin2026verification}
Ruinan Jin, Beidi Zhao, Myeongkyun Kang, Qiong Zhang, and Xiaoxiao Li.
\newblock Verification mirage: Mapping the reliability boundary of
  self-verification in medical {VQA}.
\newblock \emph{arXiv preprint arXiv:2605.10850}, 2026.

\bibitem[Katz et~al.(2024)Katz, Cohen, Shachar, Somer, Fink, Morse, Shreiber,
  and Wolf]{katz2024gpt}
Uriel Katz, Eran Cohen, Eliya Shachar, Jonathan Somer, Adam Fink, Eli Morse,
  Beki Shreiber, and Ido Wolf.
\newblock {GPT} versus resident physicians --- a benchmark based on official
  board scores.
\newblock \emph{NEJM AI}, 1\penalty0 (5):\penalty0 AIdbp2300192, 2024.

\bibitem[Kermany et~al.(2018)Kermany, Goldbaum, Cai, Valentim, Liang, Baxter,
  McKeown, Yang, Wu, Yan, et~al.]{kermany2018identifying}
Daniel~S. Kermany, Michael Goldbaum, Wenjia Cai, Carolina C.~S. Valentim,
  Huiying Liang, Sally~L. Baxter, Alex McKeown, Ge Yang, Xiaokang Wu, Fangbing
  Yan, et~al.
\newblock Identifying medical diagnoses and treatable diseases by image-based
  deep learning.
\newblock \emph{Cell}, 172\penalty0 (5):\penalty0 1122--1131.e9, 2018.

\bibitem[Kim et~al.(2024)Kim, Park, Jeong, Chan, Xu, McDuff, Lee, Ghassemi,
  Breazeal, and Park]{kim2024mdagents}
Yubin Kim, Chanwoo Park, Hyewon Jeong, Yik~Siu Chan, Xuhai Xu, Daniel McDuff,
  Hyeonhoon Lee, Marzyeh Ghassemi, Cynthia Breazeal, and Hae~Won Park.
\newblock {MDAgents}: An adaptive collaboration of {LLMs} for medical
  decision-making.
\newblock In \emph{Advances in Neural Information Processing Systems}, pages
  79410--79452, 2024.

\bibitem[Kirillov et~al.(2023)Kirillov, Mintun, Ravi, Mao, Rolland, Gustafson,
  Xiao, Whitehead, Berg, Lo, Doll{\'a}r, and Girshick]{kirillov2023segment}
Alexander Kirillov, Eric Mintun, Nikhila Ravi, Hanzi Mao, Chloe Rolland, Laura
  Gustafson, Tete Xiao, Spencer Whitehead, Alexander~C. Berg, Wan-Yen Lo, Piotr
  Doll{\'a}r, and Ross Girshick.
\newblock Segment anything.
\newblock In \emph{Proceedings of the IEEE/CVF International Conference on
  Computer Vision (ICCV)}, pages 3992--4003, 2023.

\bibitem[Lewis et~al.(2020)Lewis, Perez, Piktus, Petroni, Karpukhin, Goyal,
  K{\"u}ttler, Lewis, Yih, Rockt{\"a}schel, Riedel, and
  Kiela]{lewis2020retrieval}
Patrick Lewis, Ethan Perez, Aleksandra Piktus, Fabio Petroni, Vladimir
  Karpukhin, Naman Goyal, Heinrich K{\"u}ttler, Mike Lewis, Wen-tau Yih, Tim
  Rockt{\"a}schel, Sebastian Riedel, and Douwe Kiela.
\newblock Retrieval-augmented generation for knowledge-intensive {NLP} tasks.
\newblock In \emph{Advances in Neural Information Processing Systems
  (NeurIPS)}, pages 9459--9474, 2020.

\bibitem[Li et~al.(2024)Li, Yan, Pan, Luo, Ji, Ding, Xu, Liu, Dong, Lin, and
  Wang]{li2024mmedagent}
Binxu Li, Tiankai Yan, Yuanting Pan, Jie Luo, Ruiyang Ji, Jiayuan Ding, Zhe Xu,
  Shilong Liu, Haoyu Dong, Zihao Lin, and Yixin Wang.
\newblock {MMedAgent}: Learning to use medical tools with multi-modal agent.
\newblock In \emph{Findings of the Association for Computational Linguistics:
  EMNLP 2024}, pages 8745--8760, Miami, Florida, USA, 2024. Association for
  Computational Linguistics.

\bibitem[Li et~al.(2025)Li, Jiang, Xu, Xia, Jia, Chen, and Shi]{li2025atcxr}
Xueyang Li, Mingze Jiang, Gelei Xu, Jun Xia, Mengzhao Jia, Danny Chen, and Yiyu
  Shi.
\newblock {AT-CXR}: Uncertainty-aware agentic triage for chest {X}-rays.
\newblock \emph{arXiv preprint arXiv:2508.19322}, 2025.

\bibitem[Liu et~al.(2024)Liu, Lin, Hewitt, Paranjape, Bevilacqua, Petroni, and
  Liang]{liu2024lost}
Nelson~F. Liu, Kevin Lin, John Hewitt, Ashwin Paranjape, Michele Bevilacqua,
  Fabio Petroni, and Percy Liang.
\newblock Lost in the middle: How language models use long contexts.
\newblock \emph{Transactions of the Association for Computational Linguistics},
  12:\penalty0 157--173, 2024.

\bibitem[Lou et~al.(2025)Lou, Yang, Yu, Fu, Han, Huang, and
  Yu]{lou2025cxragent}
Jinhui Lou, Yan Yang, Zhou Yu, Zhenqi Fu, Weidong Han, Qingming Huang, and Jun
  Yu.
\newblock {CXRAgent}: Director-orchestrated multi-stage reasoning for chest
  x-ray interpretation.
\newblock \emph{arXiv preprint arXiv:2510.21324}, 2025.

\bibitem[Lu et~al.(2026)Lu, Liang, Pan, Zhang, Dong, Leng, Cui, Wu, and
  Zhang]{lu2026medr2}
Keer Lu, Zheng Liang, Da Pan, Shusen Zhang, Guosheng Dong, Huang Leng, Bin Cui,
  Zhonghai Wu, and Wentao Zhang.
\newblock {Med-R}$^2$: Crafting trustworthy {LLM} physicians via retrieval and
  reasoning of evidence-based medicine.
\newblock In \emph{Proceedings of the ACM Web Conference 2026}, pages
  1864--1875. ACM, 2026.

\bibitem[Ma et~al.(2024)Ma, He, Li, Han, You, and Wang]{ma2024segment}
Jun Ma, Yuting He, Feifei Li, Lin Han, Chenyu You, and Bo Wang.
\newblock Segment anything in medical images.
\newblock \emph{Nature Communications}, 15\penalty0 (1):\penalty0 654, 2024.

\bibitem[Ma et~al.(2026)Ma, Yu, Qiu, Yang, An, Dong, Zhao, and
  Liu]{ma2026activemedagent}
Weiwei Ma, Xiaobing Yu, Peijie Qiu, Jin Yang, Zhaoqi An, Xuanzhao Dong, Xiaoqi
  Zhao, and Xiaofeng Liu.
\newblock {ActiveMedAgent}: Cost-aware trajectory learning for multimodal
  medical diagnosis.
\newblock In \emph{Findings of the Association for Computational Linguistics:
  EMNLP 2026}, pages 30576--30595, Budapest, Hungary, 2026. Association for
  Computational Linguistics.

\bibitem[Messina et~al.(2026)Messina, Villa, Le{\'o}n~Alc{\'a}zar, S{\'a}nchez,
  Hinojosa, Parra, Soto, and Ghanem]{messina2026cure}
Pablo Messina, Andr{\'e}s Villa, Juan Le{\'o}n~Alc{\'a}zar, Karen S{\'a}nchez,
  Carlos Hinojosa, Denis Parra, {\'A}lvaro Soto, and Bernard Ghanem.
\newblock {CURE}: Curriculum-guided multi-task training for reliable anatomy
  grounded report generation.
\newblock In \emph{Proceedings of the IEEE/CVF Conference on Computer Vision
  and Pattern Recognition (CVPR)}, pages 36279--36289, 2026.

\bibitem[M{\"u}ller et~al.(2024)M{\"u}ller, Kaissis, and
  Rueckert]{muller2024chex}
Philip M{\"u}ller, Georgios Kaissis, and Daniel Rueckert.
\newblock {ChEX}: Interactive localization and region description in chest
  {X}-rays.
\newblock In \emph{Computer Vision -- {ECCV} 2024}, pages 92--111. Springer
  Nature Switzerland, 2024.

\bibitem[Pal et~al.(2022)Pal, Umapathi, and Sankarasubbu]{pal2022medmcqa}
Ankit Pal, Logesh~Kumar Umapathi, and Malaikannan Sankarasubbu.
\newblock {MedMCQA}: A large-scale multi-subject multi-choice dataset for
  medical domain question answering.
\newblock In \emph{Proceedings of the Conference on Health, Inference, and
  Learning}, pages 248--260. PMLR, 2022.

\bibitem[P{\'e}rez-Garc{\'i}a et~al.(2025)P{\'e}rez-Garc{\'i}a, Sharma,
  Bond-Taylor, Bouzid, Salvatelli, Ilse, Bannur, Castro, Schwaighofer, Lungren,
  Wetscherek, Codella, Hyland, Alvarez-Valle, and
  Oktay]{perezgarcia2025exploring}
Fernando P{\'e}rez-Garc{\'i}a, Harshita Sharma, Sam Bond-Taylor, Kenza Bouzid,
  Valentina Salvatelli, Maximilian Ilse, Shruthi Bannur, Daniel~C. Castro,
  Anton Schwaighofer, Matthew~P. Lungren, Maria~Teodora Wetscherek, Noel
  Codella, Stephanie~L. Hyland, Javier Alvarez-Valle, and Ozan Oktay.
\newblock Exploring scalable medical image encoders beyond text supervision.
\newblock \emph{Nature Machine Intelligence}, 7\penalty0 (1):\penalty0
  119--130, 2025.

\bibitem[Raiaan and Fahad(2026)]{raiaan2026evade}
Mohaimenul Azam~Khan Raiaan and Nur~Mohammad Fahad.
\newblock {EVADE}: Evidence-verified agentic diagnosis with escape.
\newblock \emph{arXiv preprint arXiv:2608.18833}, 2026.

\bibitem[Robertson and Zaragoza(2009)]{robertson2009probabilistic}
Stephen Robertson and Hugo Zaragoza.
\newblock The probabilistic relevance framework: {BM25} and beyond.
\newblock \emph{Foundations and Trends in Information Retrieval}, 3\penalty0
  (4):\penalty0 333--389, 2009.

\bibitem[Sellergren et~al.(2026)Sellergren, Kazemzadeh, Mahvar, Kiraly,
  Traverse, Kohlberger, Gao, Jaroensri, Xu, Jamil, et~al.]{sellergren2026open}
Andrew Sellergren, Sahar Kazemzadeh, Fereshteh Mahvar, Atilla Kiraly, Madeleine
  Traverse, Timo Kohlberger, Chufan Gao, Ronnachai Jaroensri, Shawn Xu, Fayaz
  Jamil, et~al.
\newblock An open vision-language model for diverse medical applications.
\newblock \emph{Nature Medicine}, 32\penalty0 (10):\penalty0 3695--3704, 2026.

\bibitem[Shen et~al.(2026)Shen, Jian, Li, Liu, Moll, Hu, Rueckert, Li, and
  Pan]{shen2026evomedagent}
Weixiang Shen, Bailiang Jian, Jun Li, Che Liu, Johannes Moll, Xiaobin Hu,
  Daniel Rueckert, Hongwei~Bran Li, and Jiazhen Pan.
\newblock {Evo-MedAgent}: Beyond one-shot diagnosis with agents that remember,
  reflect, and improve.
\newblock \emph{arXiv preprint arXiv:2604.14475}, 2026.

\bibitem[Shi et~al.(2023)Shi, Chen, Misra, Scales, Dohan, Chi, Sch{\"a}rli, and
  Zhou]{shi2023large}
Freda Shi, Xinyun Chen, Kanishka Misra, Nathan Scales, David Dohan, Ed~H. Chi,
  Nathanael Sch{\"a}rli, and Denny Zhou.
\newblock Large language models can be easily distracted by irrelevant context.
\newblock In \emph{Proceedings of the 40th International Conference on Machine
  Learning (ICML)}, pages 31210--31227. PMLR, 2023.

\bibitem[Tang et~al.(2024)Tang, Zou, Zhang, Li, Zhao, Zhang, Cohan, and
  Gerstein]{tang2024medagents}
Xiangru Tang, Anni Zou, Zhuosheng Zhang, Ziming Li, Yilun Zhao, Xingyao Zhang,
  Arman Cohan, and Mark Gerstein.
\newblock {MedAgents}: Large language models as collaborators for zero-shot
  medical reasoning.
\newblock In \emph{Findings of the Association for Computational Linguistics:
  ACL 2024}, pages 599--621, Bangkok, Thailand, 2024. Association for
  Computational Linguistics.

\bibitem[Wang et~al.(2026{\natexlab{a}})Wang, Yang, Wu, Ji, Ye, Chen, Xiong,
  Fang, Liu, Xu, Shan, Liu, He, and Liu]{wang2026mira}
Shengzhi Wang, Jun Yang, Kai Wu, Xiaozhong Ji, Yiwen Ye, Ziyang Chen, Mingliang
  Xiong, Wen Fang, Mingqing Liu, Mengyuan Xu, Miaoxuan Shan, Caiyan Liu, Bin
  He, and Qingwen Liu.
\newblock {MIRA}: Medical image reflection for agentic diagnosis.
\newblock \emph{arXiv preprint arXiv:2608.10827}, 2026{\natexlab{a}}.

\bibitem[Wang et~al.(2023)Wang, Wei, Schuurmans, Le, Chi, Narang, Chowdhery,
  and Zhou]{wang2023selfconsistency}
Xuezhi Wang, Jason Wei, Dale Schuurmans, Quoc~V. Le, Ed~H. Chi, Sharan Narang,
  Aakanksha Chowdhery, and Denny Zhou.
\newblock Self-consistency improves chain of thought reasoning in language
  models.
\newblock In \emph{The Eleventh International Conference on Learning
  Representations (ICLR)}, 2023.

\bibitem[Wang et~al.(2026{\natexlab{b}})Wang, Wu, Cai, Low, Yang, Li, and
  Jin]{wang2026medagentpro}
Ziyue Wang, Junde Wu, Linghan Cai, Chang~Han Low, Xihong Yang, Qiaxuan Li, and
  Yueming Jin.
\newblock {MedAgent-Pro}: Towards evidence-based multi-modal medical diagnosis
  via reasoning agentic workflow.
\newblock In \emph{The Fourteenth International Conference on Learning
  Representations}, 2026{\natexlab{b}}.

\bibitem[Wu et~al.(2026)Wu, Tang, Li, Kai, Yuan, Sun, Chen, and
  Wang]{wu2026from}
Wenhao Wu, Zhentao Tang, Yafu Li, Shixiong Kai, Mingxuan Yuan, Zhenhong Sun,
  Chunlin Chen, and Zhi Wang.
\newblock From conflict to consensus: Boosting medical reasoning via
  multi-round agentic {RAG}.
\newblock In \emph{Forty-third International Conference on Machine Learning
  (ICML)}, 2026.

\bibitem[Xie and Zhang(2025)]{xie2025cureagent}
Ting-Ting Xie and Yixin Zhang.
\newblock {CureAgent}: A training-free executor-analyst framework for clinical
  reasoning.
\newblock \emph{arXiv preprint arXiv:2512.05576}, 2025.

\bibitem[Xiong et~al.(2024)Xiong, Jin, Lu, and Zhang]{xiong2024benchmarking}
Guangzhi Xiong, Qiao Jin, Zhiyong Lu, and Aidong Zhang.
\newblock Benchmarking retrieval-augmented generation for medicine.
\newblock In \emph{Findings of the Association for Computational Linguistics:
  ACL 2024}, pages 6233--6251, Bangkok, Thailand, 2024. Association for
  Computational Linguistics.

\bibitem[Xu et~al.(2026)Xu, Chan, Li, Aljunied, Yuan, Wang, Xiao, Chen, Liu,
  Li, et~al.]{xu2026lingshu}
Weiwen Xu, Hou~Pong Chan, Long Li, Mahani Aljunied, Ruifeng Yuan, Jianyu Wang,
  Chenghao Xiao, Guizhen Chen, Chaoqun Liu, Zhaodonghui Li, et~al.
\newblock Lingshu: Generalist foundation model for unified multimodal medical
  understanding and reasoning.
\newblock \emph{IEEE Transactions on Pattern Analysis and Machine
  Intelligence}, 2026.

\bibitem[Xu et~al.(2020)Xu, Zhao, Song, Stewart, and Ermon]{xu2020theory}
Yilun Xu, Shengjia Zhao, Jiaming Song, Russell Stewart, and Stefano Ermon.
\newblock A theory of usable information under computational constraints.
\newblock In \emph{International Conference on Learning Representations
  (ICLR)}, 2020.

\bibitem[Yan et~al.(2024)Yan, Gu, Zhu, and Ling]{yan2024corrective}
Shi-Qi Yan, Jia-Chen Gu, Yun Zhu, and Zhen-Hua Ling.
\newblock Corrective retrieval augmented generation.
\newblock \emph{arXiv preprint arXiv:2401.15884}, 2024.

\bibitem[Yang et~al.(2025)Yang, Li, Yang, Zhang, Hui, Zheng, Yu, Gao, Huang,
  Lv, et~al.]{yang2025qwen3}
An Yang, Anfeng Li, Baosong Yang, Beichen Zhang, Binyuan Hui, Bo Zheng, Bowen
  Yu, Chang Gao, Chengen Huang, Chenxu Lv, et~al.
\newblock {Qwen3} technical report.
\newblock \emph{arXiv preprint arXiv:2505.09388}, 2025.

\bibitem[Yao et~al.(2023)Yao, Zhao, Yu, Du, Shafran, Narasimhan, and
  Cao]{yao2023react}
Shunyu Yao, Jeffrey Zhao, Dian Yu, Nan Du, Izhak Shafran, Karthik Narasimhan,
  and Yuan Cao.
\newblock {ReAct}: Synergizing reasoning and acting in language models.
\newblock In \emph{The Eleventh International Conference on Learning
  Representations (ICLR)}, 2023.

\bibitem[Zhang et~al.(2026)Zhang, Barrett, Kim, Sun, Taghavi, and
  Kenthapadi]{zhang2026radagents}
Kai Zhang, Corey~D. Barrett, Jangwon Kim, Lichao Sun, Tara Taghavi, and
  Krishnaram Kenthapadi.
\newblock {RadAgents}: Multimodal agentic reasoning for chest x-ray
  interpretation with radiologist-like workflows.
\newblock In \emph{Proceedings of The 9th International Conference on Medical
  Imaging with Deep Learning}, pages 3496--3519. PMLR, 2026.

\bibitem[Zhang et~al.(2025)Zhang, Xu, Usuyama, Xu, Bagga, Tinn, Preston, Rao,
  Wei, Valluri, et~al.]{zhang2025multimodal}
Sheng Zhang, Yanbo Xu, Naoto Usuyama, Hanwen Xu, Jaspreet Bagga, Robert Tinn,
  Sam Preston, Rajesh Rao, Mu Wei, Naveen Valluri, et~al.
\newblock A multimodal biomedical foundation model trained from fifteen million
  image--text pairs.
\newblock \emph{NEJM AI}, 2\penalty0 (1):\penalty0 AIoa2400640, 2025.

\bibitem[Zhao et~al.(2025)Zhao, Gu, Yang, Usuyama, Lee, Kiblawi, Naumann, Gao,
  Crabtree, Abel, et~al.]{zhao2025foundation}
Theodore Zhao, Yu Gu, Jianwei Yang, Naoto Usuyama, Ho~Hin Lee, Sid Kiblawi,
  Tristan Naumann, Jianfeng Gao, Angela Crabtree, Jacob Abel, et~al.
\newblock A foundation model for joint segmentation, detection and recognition
  of biomedical objects across nine modalities.
\newblock \emph{Nature Methods}, 22\penalty0 (1):\penalty0 166--176, 2025.

\bibitem[Zhou et~al.(2025)Zhou, Song, and Shen]{zhou2025mam}
Yucheng Zhou, Lingran Song, and Jianbing Shen.
\newblock {MAM}: Modular multi-agent framework for multi-modal medical
  diagnosis via role-specialized collaboration.
\newblock In \emph{Findings of the Association for Computational Linguistics:
  ACL 2025}, pages 25319--25333, Vienna, Austria, 2025. Association for
  Computational Linguistics.

\bibitem[Zhu et~al.(2025)Zhu, Wang, Chen, Liu, Ye, Gu, Tian, Duan, Su, Shao,
  et~al.]{zhu2025internvl3}
Jinguo Zhu, Weiyun Wang, Zhe Chen, Zhaoyang Liu, Shenglong Ye, Lixin Gu, Hao
  Tian, Yuchen Duan, Weijie Su, Jie Shao, et~al.
\newblock {InternVL3}: Exploring advanced training and test-time recipes for
  open-source multimodal models.
\newblock \emph{arXiv preprint arXiv:2504.10479}, 2025.

\end{thebibliography}
}

\end{document}